\documentclass[10pt,twocolumn,letterpaper]{article}

\usepackage[pagenumbers]{cvpr} % To force page numbers, e.g. for an arXiv version

\definecolor{cvprblue}{rgb}{0.21,0.49,0.74}
\usepackage[pagebackref,breaklinks,colorlinks,allcolors=cvprblue]{hyperref}
\usepackage{algorithm}
\usepackage{algpseudocode}
\usepackage{amsmath} % Required for \text, \arg\max, and math symbols
\usepackage{booktabs}
\usepackage{multirow}
\usepackage{graphicx}
\usepackage[multiple]{footmisc}
\usepackage{bigfoot}
\usepackage{bbding} % For the \Envelope symbol

\def\paperID{7} % *** Enter the Paper ID here
\def\confName{CVPR}
\def\confYear{2026}

\title{EdgeDiT: Hardware-Aware Diffusion Transformers for Efficient On-Device Image Generation}

\renewcommand{\footnoterule}{\vfill\kern -3pt \hrule width \linewidth \kern 2.6pt}

\author{
    Sravanth Kodavanti\thanks{%
        \begin{tabular}[t]{@{}l@{}}
            Equal contribution. \\
            Email: {\tt \{k.sravanth, at.manjunath\}@samsung.com}
        \end{tabular}} \quad
    Manjunath Arveti\footnotemark[1] \quad
    Sowmya Vajrala \quad
    Srinivas Miriyala \quad
    Vikram N R \\[0.5ex]
    Samsung Research Institute Bangalore, India 
}

\begin{document}
\maketitle
\begin{abstract}
Diffusion Transformers (DiT) have established a new state-of-the-art in high-fidelity image synthesis; however, their massive computational complexity and memory requirements hinder local deployment on resource-constrained edge devices. In this paper, we introduce EdgeDiT, a family of hardware-efficient generative transformers specifically engineered for mobile Neural Processing Units (NPUs), such as the Qualcomm Hexagon and Apple Neural Engine (ANE). By leveraging a hardware-aware optimization framework, we systematically identify and prune structural redundancies within the DiT backbone that are particularly taxing for mobile data-flows. Our approach yields a series of lightweight models that achieve a 20–30\% reduction in parameters, a 36-46\% decrease in FLOPs, and a 1.65-fold reduction in on-device latency without sacrificing the scaling advantages or the expressive capacity of the original transformer architecture. Extensive benchmarking demonstrates that EdgeDiT offers a superior Pareto-optimal trade-off between Fréchet Inception Distance (FID) and inference latency compared to both optimized mobile U-Nets and vanilla DiT variants. By enabling responsive, private, and offline generative AI directly on-device, EdgeDiT provides a scalable blueprint for transitioning large-scale foundation models from high-end GPUs to the palm of the user.
\end{abstract}
% \begin{figure}
%     \centering
%     \includegraphics[width=0.5\linewidth]{image.png}
%     \caption{Enter Caption}
%     \label{fig:placeholder}
% \end{figure}    
\section{Introduction}

Diffusion models have recently emerged as a dominant paradigm for high-fidelity image generation. Early approaches such as Denoising Diffusion Probabilistic Models (DDPM) \cite{ho2020ddpm} and Denoising Diffusion Implicit Models (DDIM) \cite{song2021ddim} demonstrated that iterative denoising processes can generate high-quality images by gradually transforming Gaussian noise into structured samples. These models were initially built upon convolutional U-Net backbones, which became the de facto architecture for diffusion-based image synthesis.

Recent advances have shifted toward transformer-based generative architectures. Diffusion Transformers (DiT) \cite{peebles2023dit} replace convolutional U-Nets with Vision Transformer backbones, enabling improved scalability and better utilization of large-scale training data. Building on this paradigm, several works have explored architectural improvements and training strategies for transformer-based diffusion models. Masked Diffusion Transformer (MDT) \cite{gao2023mdt} introduced masked modeling objectives that significantly improve synthesis quality. Scalable Interpolant Transformers (SiT) \cite{siT2024} further unified diffusion and flow-based generative modeling under a transformer-based framework. More recently, representation alignment techniques \cite{li2024representation} have demonstrated that training DiTs can be simplified and stabilized through improved feature alignment strategies.

Parallel efforts have also investigated alternative backbones for diffusion models. For instance, state-space models have recently been explored as efficient sequence modeling alternatives to transformers, leading to architectures such as diffusion models with state-space backbones \cite{statespace2024diffusion}. These approaches aim to improve scalability and computational efficiency while preserving generative quality.

Despite these advances, the deployment of diffusion transformers on edge devices remains challenging. State-of-the-art DiT architectures require substantial computational resources, large memory footprints, and high inference latency, making them impractical for mobile and embedded platforms. While cloud-based inference mitigates these limitations, it introduces privacy concerns, network dependency, and increased energy consumption.

In this work, we address the challenge of bringing transformer-based diffusion models to resource-constrained hardware. We introduce \textbf{EdgeDiT}, a family of lightweight diffusion transformers specifically optimized for mobile Neural Processing Units (NPUs) such as Qualcomm NPUs and Apple Neural Engine (ANE). Our approach identifies the computationally expensive and redundant operations in the pretrained DiT model and applies hardware-aware architectural optimization to improve on-device inference efficiency, while preserving generative capacity.

Through systematic model redesign and pruning strategies, EdgeDiT achieves a 20--30\% reduction in parameters, a 36--46\% decrease in FLOPs, and 1.65$\times$ speedup in on-device latency compared to baseline DiT architectures. Extensive experiments demonstrate that EdgeDiT achieves competitive Fréchet Inception Distance (FID) while significantly reducing inference latency on edge hardware.

Our contributions are summarized as follows:

\begin{itemize}
\item We propose EdgeDiT, a hardware-aware diffusion transformer architecture designed for efficient on-device generation.
\item We introduce structural simplifications and pruning strategies tailored for mobile NPUs.
\item We demonstrate improved Pareto trade-offs between generation quality and inference efficiency compared to existing DiT variants.
\item We provide empirical evidence that transformer-based diffusion models can be effectively scaled down for real-world edge deployment.
\end{itemize}

\section{Related Work}

\subsection{Diffusion Models}

Diffusion models have achieved remarkable success in generative modeling across a variety of visual tasks. Denoising Diffusion Probabilistic Models (DDPM)~\cite{ho2020ddpm} introduced a probabilistic formulation for image generation based on iterative denoising of Gaussian noise. Subsequent improvements such as Improved DDPM~\cite{nichol2021improved} enhanced sampling efficiency and training stability. Latent Diffusion Models (LDM)~\cite{rombach2022ldm} further improved scalability by performing diffusion in a compressed latent space, significantly reducing computational requirements while maintaining high visual fidelity.

\subsection{Diffusion Transformers}

Transformer architectures have recently been adopted as powerful backbones for diffusion models. Diffusion Transformers (DiT)~\cite{peebles2023dit} replace convolutional U-Net architectures with transformer blocks operating on latent tokens, demonstrating strong scaling properties for generative modeling. Several works have further explored architectural improvements to diffusion transformers. Masked Diffusion Transformer (MDT) \cite{gao2023mdt} and its improved variant MDTv2~\cite{gao2023mdtv2} introduce masked latent modeling strategies that improve contextual reasoning and accelerate training. 

Scalable Interpolant Transformers (SiT)~\cite{siT2024} extend diffusion transformers to a broader generative modeling framework that unifies diffusion and flow-based models through stochastic interpolants, enabling flexible training objectives and improved performance across model scales.

Other recent work investigates alternative architectural backbones for diffusion models. For instance, state-space model based diffusion architectures replace attention mechanisms with structured state-space layers to improve scalability and reduce computational complexity. These approaches highlight ongoing efforts to improve both the scalability and efficiency of transformer-based diffusion models.

\subsection{Efficient and Lightweight Diffusion Models}

The growing computational cost of diffusion models has motivated research into lightweight diffusion architectures. MobileDiffusion~\cite{mobilediffusion} proposes architectural simplifications designed for mobile hardware to enable efficient on-device diffusion inference. Other works explore structured pruning and token sparsification techniques to reduce redundancy in diffusion transformers. 

In addition, efficient training strategies have been explored to reduce the cost of training large diffusion models. For example, masked training strategies and representation alignment approaches have been shown to significantly simplify the diffusion transformer training process while maintaining strong generative performance~\cite{li2024representation}.

\subsection{Model Compression and Architecture Search}

Model compression techniques such as pruning, knowledge distillation, and neural architecture search have been widely used to obtain compact neural networks suitable for deployment on edge hardware. Knowledge distillation~\cite{hinton2015distilling} allows smaller student networks to mimic the behavior of larger teacher models while preserving performance. Automated architecture search methods such as Bayesian optimization~\cite{snoek2012practical} enable systematic exploration of large architecture spaces under hardware constraints.

In contrast to prior work that primarily focuses on manual architecture design or pruning strategies, our work introduces a surrogate-based architecture search framework that combines feature-wise knowledge distillation and multi-objective Bayesian optimization to discover efficient diffusion transformer architectures optimized for edge deployment.

\section{Method}

\begin{figure*}
    \centering
    \includegraphics[width=1.0\linewidth]{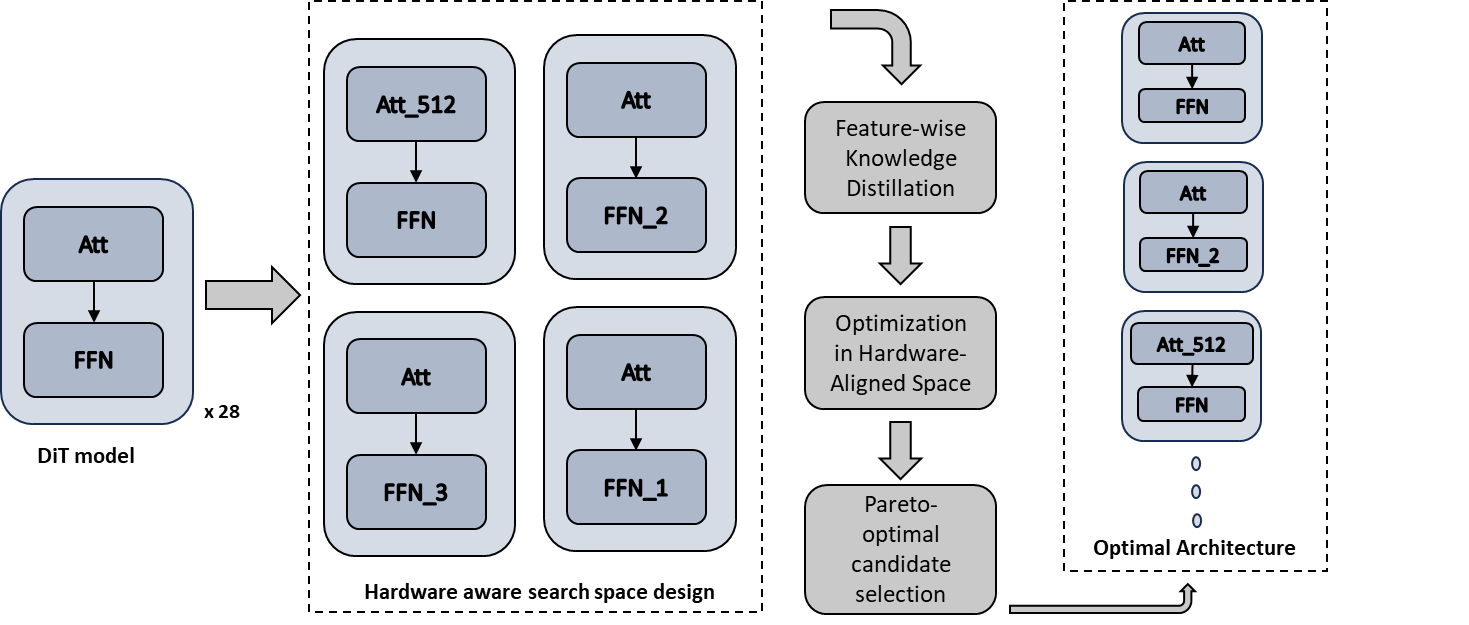}
    % \caption{Surrogates design}
    \caption{Flowsheet depicting the proposed method}
    \label{fig:flow}
\end{figure*}

% Our goal is to design hardware-efficient variants of Diffusion Transformers that preserve generative performance while reducing computational and memory overhead for edge deployment. Starting from a pretrained DiT-XL/2 backbone, we construct a structured architecture search space consisting of lightweight surrogate transformer blocks. These surrogates are trained using feature-wise knowledge distillation and combined to form candidate architectures. We then employ multi-objective Bayesian optimization to identify Pareto-optimal models in the hardware–performance space.

Our goal is to design edge-efficient variants of Diffusion Transformers based on the DiT-XL/2~\cite{peebles2023dit} baseline, that preserve generative performance while reducing the computational and memory overhead for edge deployment. Moreover, theoretical compute metrics (FLOPs/GMACs) may not always reliably predict pragmatic latency, since NPUs are optimized for specific operations such as GEMM. So, reducing the arithmetic compute may not always guarantee proportional edge latency improvements. To address this gap, we construct a structured architecture search space of lightweight surrogate transformer blocks, which are aligned with edge hardware. 

These surrogates are trained using feature-wise knowledge distillation and these locally distilled surrogates are combined to form candidate architectures, which are then fully trained end-to-end for accuracy improvements. We employ multi-objective Bayesian optimization to identify these Pareto-optimal candidate models in the hardware-performance search space. From this set, we select the most balanced architecture and fix it as our base architecture \textit{EdgeDiT}. An overview of the proposed framework is illustrated in Figure \ref{fig:flow}.

\subsection{Hardware-aware surgery of DiTs}

Diffusion Transformers (DiT)~\cite{peebles2023dit} operate on latent image representations by treating them as sequences of tokens processed by transformer blocks. Given an input image $x$, it is first encoded into a latent representation using a variational autoencoder. The latent representation is then divided into non-overlapping patches of size $p \times p$, producing a sequence of tokens that form the input to the transformer backbone. A DiT model consists of a stack of $L$ ($L=28$ for DiT-XL/2 \cite{peebles2023dit}) transformer blocks. Each block contains two primary components:

\begin{itemize}
\item Multi-head self-attention (MHSA)
\item A feed-forward network (FFN) with an expansion ratio $r$
\end{itemize}
Formally, a transformer block can be expressed as

\begin{equation}
h_{l+1} = h_l + \text{MHSA}(\text{LN}(h_l)) + \text{FFN}(\text{LN}(h_l)),
\end{equation}

where $h_l$ denotes the hidden representation at layer $l$ and LN denotes layer normalization.

While DiT models provide strong generative performance, their large number of layers and high-dimensional hidden representations introduce significant computational and memory overhead. To address this challenge, we decompose the DiT architecture into smaller hardware-friendly building blocks and construct a structured architecture search space.

% Our search space consists of several architectural modifications:

% \begin{itemize}
% \item \textbf{Block removal}: Selected transformer blocks are removed to reduce network depth.
% \item \textbf{MLP ratio modification}: The expansion ratio of the feed-forward network is varied (e.g., $r \in \{1,2,3,4\}$) to reduce parameter count and FLOPs.
% \item \textbf{Hidden dimension reduction}: The internal projection dimensions of transformer layers are reduced to construct low-rank representations.
% \end{itemize}

Our hardware-aware search space consists of the following surrogates.

\begin{enumerate}
\item \textbf{Block removal}: Every two consecutive DiT layers are selected for removal to reduce the network depth.
\item \textbf{MLP ratio modification}: The expansion ratio of the FFN block in DiT layers is varied ($\text{expansion factor} = r \in \{2, 4\}$) to reduce parameter count and FLOPs.
\item \textbf{Hidden dimension reduction}: The internal projection dimension ($d \in \{512,1152\}$) of the transformer blocks in the DiT layers are reduced to construct low-rank representations.
% \item Refer to Figures \ref{fig:surr}a, \ref{fig:surr}b, \ref{fig:surr}c respectively for the surrogates list.
\end{enumerate}
The above three types of surrogates are built in two stages. 
\begin{itemize}
    \item \textbf{Stage 1} consists of choosing among the options in Block removal, which results in replacing two consecutive DiT layers with a single DiT layer, resulting in a search space of ($2 \times 2 \times ... \text{(14 times)} \times 2 = 2^{14}$) 
    \item \textbf{Stage 2} consists of choosing either the MLP modified layer or the Hidden dimension reduced layer, so per each layer we have $2 + 2 = 4$ options, resulting in a search space of $(4 \times 4 \times ... \text{(28 times)} \times 4 = 4^{28}$)
\end{itemize}

\begin{figure*}
    \centering
    \includegraphics[width=1.0\linewidth]{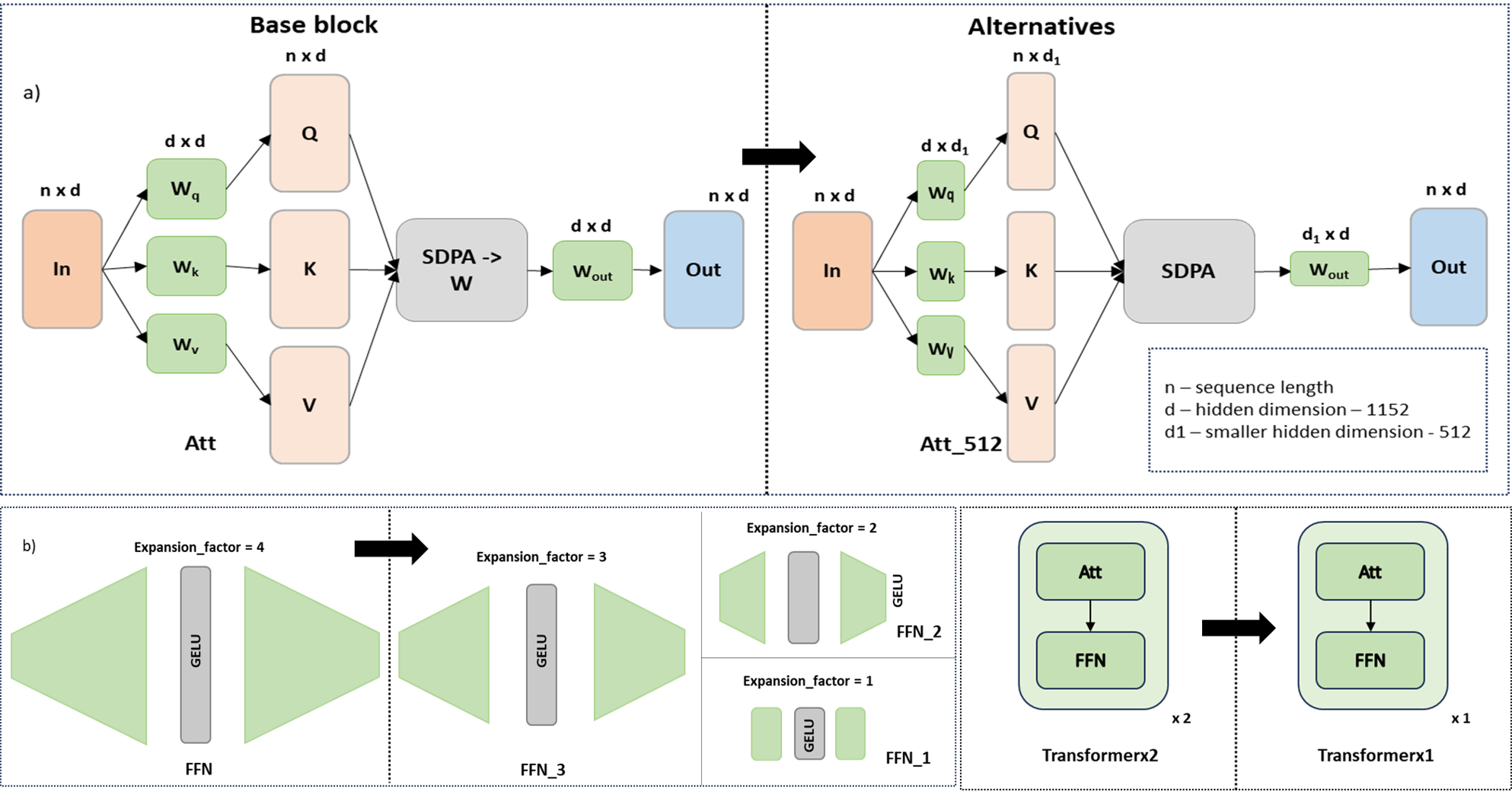}
    % \caption{Surrogates design}
    \caption{hardware-aware surrogates search space design}
    \label{fig:surr}
\end{figure*}

These modifications result in a combined hardware-aware search space of $\mathbf{2^{14} + 4^{28}}$. All possible surrogate options are listed in Figure \ref{fig:surr}.

% These modifications create a large pool of hardware-friendly surrogate blocks that can replace the original DiT blocks.

\subsection{Feature-wise Knowledge Distillation}

Training every candidate architecture in $2^{14} + 4^{28}$ from scratch would be computationally expensive and infeasible. To enable efficient training, we adopt a feature-wise knowledge distillation strategy, which is a divide-and-conquer strategy, where each surrogate block learns to mimic the behavior of the corresponding block in the teacher network.

Let $T_l(x)$ denote the output feature of the $l$-th block in the teacher model and $S_l(x)$ denote the output of the corresponding surrogate block for a given input features $x$. The feature-wise distillation loss is defined as

\begin{equation}
\mathcal{L}_{\text{KD}}^{\text{l}} = ||T_l(x) - S_l(x) ||_2^2,
\end{equation}

% where $\mathcal{S}$ denotes the set of distilled layers.

% To further stabilize training, we perform surrogate learning in two stages:

We perform surrogate training in two stages as mentioned above as well:

\textbf{Stage 1: Block removal.}
% A subset of transformer blocks is replaced with surrogate blocks while keeping the remaining blocks fixed. Each surrogate block is trained to approximate the behavior of its corresponding teacher block.
Each pair of two consecutive DiT layers are treated as the teacher model and their information is distilled into single student DiT layer.

\textbf{Stage 2: Structural Simplification.}
% In the second stage, additional structural modifications such as varying the MLP ratio and reducing hidden dimensions are introduced. Surrogate blocks are again trained using feature-wise distillation to preserve the functional characteristics of the teacher network.
In the second stage, the DiT layers act as teacher blocks and DiT layers with varying MLP ratio, reduced hidden dimensions act as students.

Since all blocks are distilled independently, this process is highly parallelizable and computationally lightweight. We obtain $14$ surrogates in stage 1, and $2 \times 28 = 56$ surrogates in stage 2. This progressive training strategy allows us to efficiently learn lightweight surrogate components, preserving DiT-XL/2's local behavior. This modular decomposition enables the design of the full candidate architecture by assembling the surrogates to approximate the DiT-XL/2 without full architecture training.
% This progressive training strategy allows us to efficiently learn lightweight surrogate components without requiring full end-to-end training for each candidate architecture.

\subsection{Assembling Candidate Architectures}

Once the surrogate blocks are trained, we construct candidate diffusion transformer architectures by combining different configurations of surrogate components.

Each candidate architecture is represented by a configuration vector

\begin{equation}
a = (b_1, b_2, \dots, b_L),
\end{equation}

where $b_i$ denotes the block type selected for layer $i$. The block type can correspond to the original DiT block or one of the surrogate variants.

Given a configuration $a$, the corresponding network is assembled by selecting the appropriate surrogate blocks for each layer. Because surrogate blocks are pre-trained through distillation, candidate architectures can be evaluated efficiently without requiring exhaustive retraining.

\subsection{Optimization in Hardware-Aligned Space}

To identify efficient architectures within the search space, we formulate model selection as a multi-objective optimization problem. Specifically, we aim to simultaneously optimize generative performance and hardware efficiency.

Let $f(a)$ denote the performance metric (e.g., FID or Inception Score) of architecture $a$, and let $g(a)$ denote a hardware cost metric such as peak memory usage or edge latency. Our objective is to identify architectures that lie on the Pareto frontier, for any architecture \textit{a} we aim to:

\begin{equation}
\max_{a \in \mathcal{A}} \; f(a) \quad \text{and} \quad \min_{a \in \mathcal{A}} \; g(a)
\end{equation}

To efficiently explore the architecture space, we employ multi-objective Bayesian optimization~\cite{snoek2012practical} (MOBO). A Gaussian model is trained to predict the objective values of candidate architectures based on previously evaluated configurations. New architectures are then selected using an acquisition function that balances exploration and exploitation.

This process enables efficient discovery of architectures that achieve favorable trade-offs between generative quality and hardware efficiency. For our experiments, we constructed a set of images from each of the candidate architectures and we took $f(a)$ as the resultant FID score and $g(a)$ as the edge latency of the particular candidate architecture.

Since the FID calculation process also remains costly even without training, we adapt MOBO whereby relaxing \textit{a} to a continuous representation $x \in [0, 1]^{28}$ and map it to the nearest feasible architecture via a deterministic mapping. Candidate architectures are selected by maximizing Expected Hyper-volume Improvement (EHVI).

% \begin{algorithm}[htbp]
% \caption{Efficient Network Search via Multi-objective Bayesian Optimization}
% \label{algo1}
% \begin{algorithmic}[1] % Number every line
% \State \textbf{Input:} Initial dataset $\mathcal{D} = \{(\mathbf{x}_i, \mathbf{y}_i)\}$ via Sobol sampling
% \State \textbf{Output:} Pareto-optimal hybrid architectures
% \While{budget not exhausted}
%     \State Fit GP surrogates for both objectives on $\mathcal{D}$
%     \State Select $\mathbf{x}^* = \arg\max \textit{EHVI}(\mathbf{x})$ using Eq.~\ref{eq2}$
%     \State Discretize $\mathbf{x}^*$ \text{to} $\mathbf{z}^*$ via stage-wise mapping
%     \State Evaluate PSNR difference and penalty for $\mathbf{z}^*$
%     \State Update $\mathcal{D} \gets \mathcal{D} \cup \{(\mathbf{x}^*, \mathbf{y}^*)\}$
% \EndWhile
% \State \Return{Non-dominated Pareto set from $\mathcal{D}$}
% \end{algorithmic}
% \end{algorithm}

\subsection{End-to-End Training}

After identifying Pareto-optimal candidate architectures from the search process, the selected models are trained end-to-end to fully adapt the network parameters. During this stage, all model components are jointly optimized using the standard diffusion training objective.

Given a noisy latent representation $z_t$ at timestep $t$, the model predicts the corresponding noise $\epsilon_\theta(z_t, t)$. The diffusion training loss is defined as

\begin{equation}
\mathcal{L}_{\text{diff}} = \mathbb{E}_{z_0, \epsilon, t}
\left[
|| \epsilon - \epsilon_\theta(z_t, t) ||_2^2
\right],
\end{equation}

where $z_0$ is the clean latent representation and $\epsilon$ is Gaussian noise.

This final training stage allows the selected architecture to fully adapt to the diffusion task, resulting in compact models that maintain strong generative performance while satisfying hardware constraints.

% \subsection{Computational Complexity Analysis}

% We further analyze the computational complexity of EdgeDiT relative to the original DiT~\cite{peebles2023dit} architecture. In transformer models, the dominant computational cost arises from the self-attention mechanism. For a sequence length $N$ and hidden dimension $d$, the complexity of a self-attention layer is $\mathcal{O}(N^2 d).$

% \begin{equation}
% \mathcal{O}(N^2 d).
% \end{equation}

% By reducing network depth, lowering hidden dimensions, and adjusting MLP expansion ratios, EdgeDiT significantly decreases both parameter count and FLOPs compared to the baseline DiT architecture. These architectural changes also reduce memory bandwidth requirements, which is a critical factor for efficient execution on mobile NPUs.

% As a result, EdgeDiT achieves substantial improvements in inference latency while maintaining competitive generative performance.
\section{Experimental Results}
\label{sec:experiments}

\subsection{Sensitivity Analysis and Surrogate choice}
\begin{figure}
    \centering
    \includegraphics[width=0.8\linewidth]{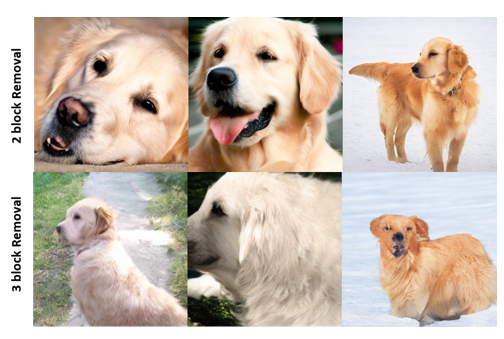}
    % \caption{Surrogates design}
    % \caption{hardware-aware surrogates search space design}
    \caption{image quality when Block Removal Surrogates are replacing 2 DiT Blocks vs 3 DiT Blocks}
    \label{fig:block_removal}
\end{figure}

We analyze the structural importance of each surrogate in the DiT architecture. We show analysis on why we chose particular surrogate instead of the other options. To validate our choices, we replace some of DiT blocks with surrogate weights learned through feature-wise-knowledge distillation approach. 
We have experimented with replacing 3 DIT Blocks with a Block removal surrogate, Figure \ref{fig:block_removal} shows that image quality drastically deteriorates when 3 DiT blocks are replaced. Due to drop in image quality, we do not include 3 DiT replacement option to our hardware-aware search space. This brought the search space from $3^{14}$ to $2^{14}$ in Stage 1.

\begin{figure}
    \centering
    \includegraphics[width=0.8\linewidth]{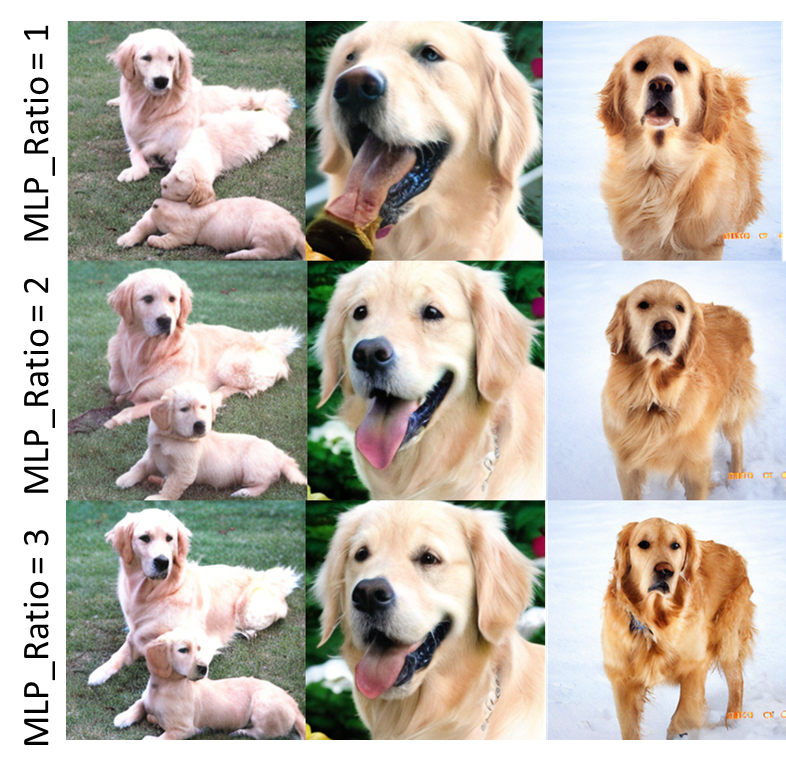}
    % \caption{Surrogates design}
    % \caption{hardware-aware surrogates search space design}
    \caption{image quality vs MLP ratio surrogates}
    \label{fig:mlp_ratio}
\end{figure}

\begin{figure}
    \centering
    \includegraphics[width=0.8\linewidth]{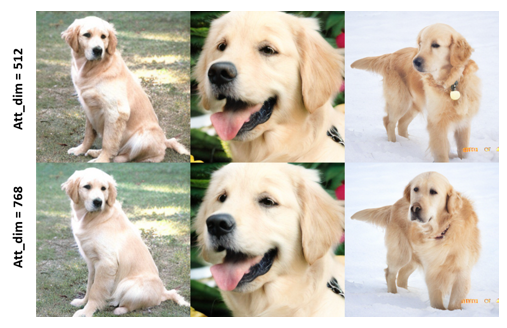}
    % \caption{Surrogates design}
    % \caption{hardware-aware surrogates search space design}
    \caption{image quality vs Hidden dimension surrogates}
    \label{fig:att_dim}
\end{figure}

\begin{figure*}
    \centering
    \includegraphics[width=1.0\linewidth]{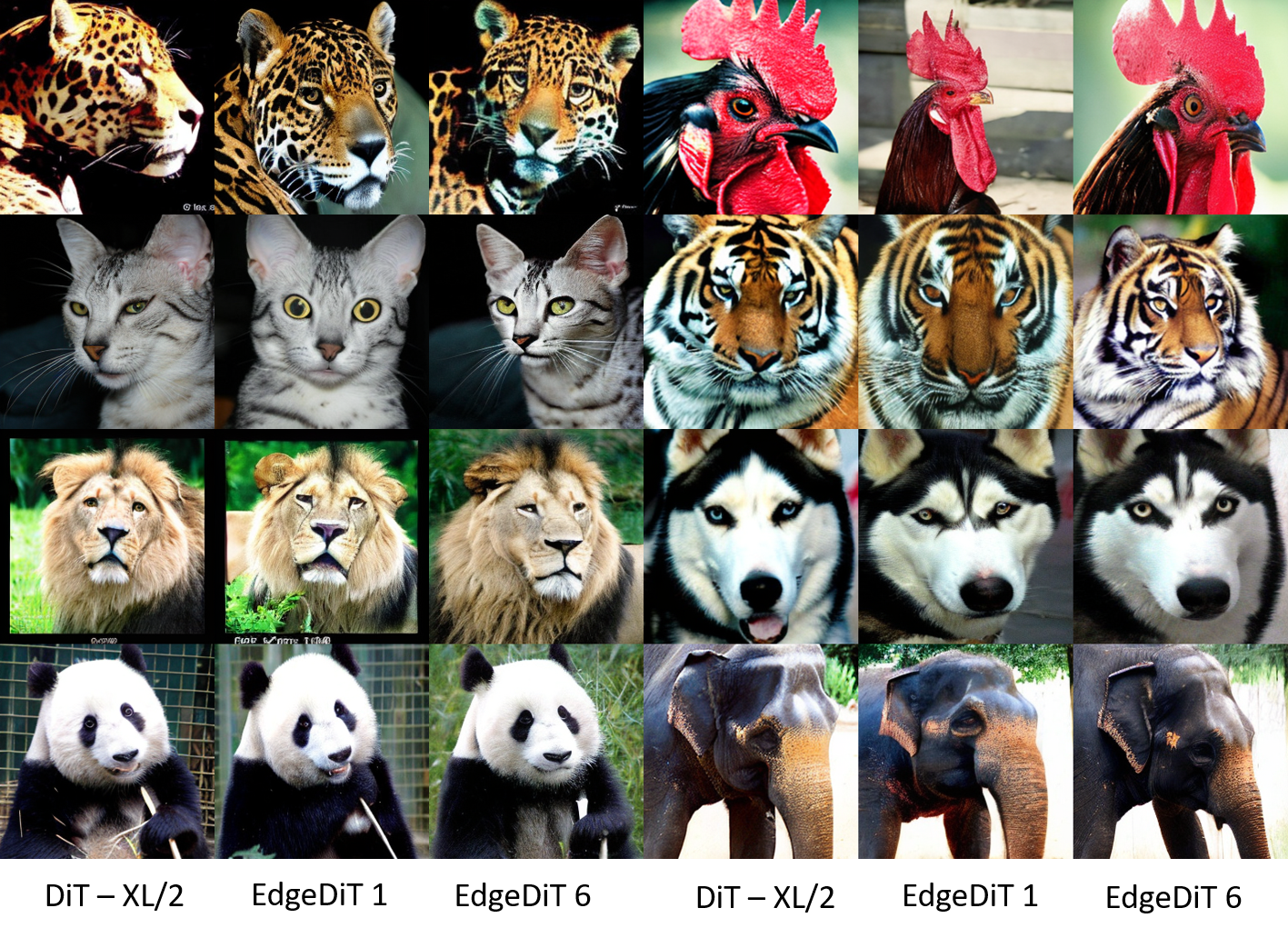}
    % \caption{Surrogates design}
    % \caption{hardware-aware surrogates search space design}
    \caption{Results of visual image quality comparison}
    \label{fig:visual_results}
\end{figure*}

\begin{figure}
    \centering
    \includegraphics[width=0.8\linewidth]{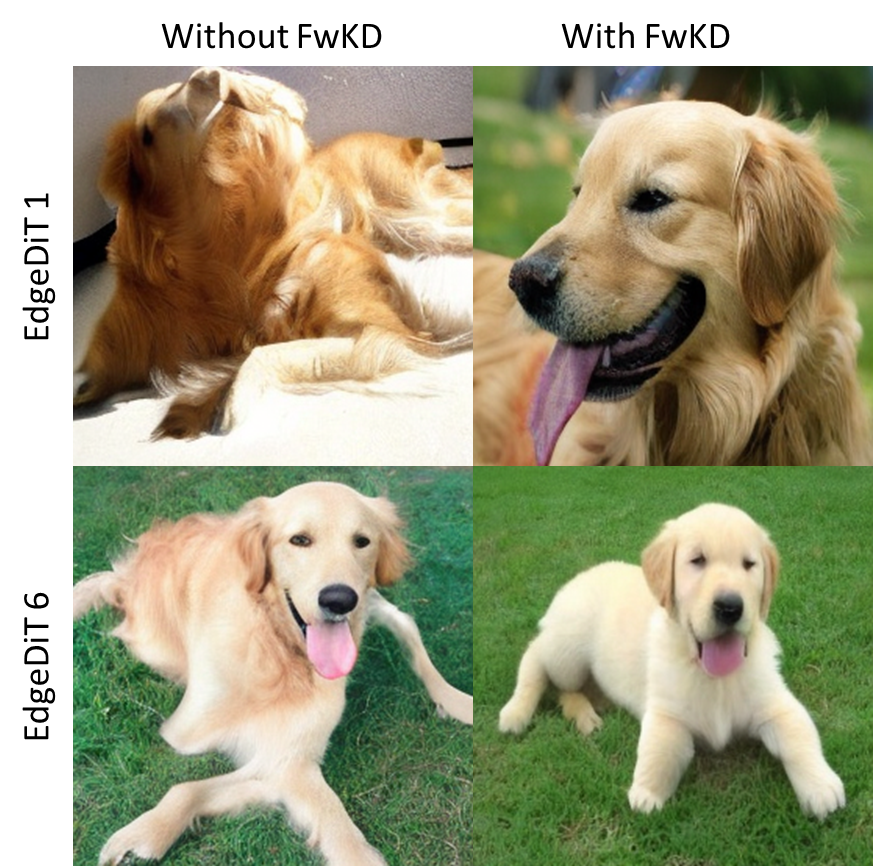}
    \caption{Ablation: image quality of EdgeDiT with and without FwKD}
    \label{fig:fwkd}
\end{figure}

\begin{table*}[]
\centering
\caption{Benchmarking class-conditional image generation on ImageNet $256 \times 256$}
\label{tab:fid_results}
\resizebox{\textwidth}{!}{%
\begin{tabular}{|cccccccc|}
\hline
\multicolumn{8}{|c|}{Class Conditioned ImageNet $256 \times 256$} \\ \hline
\multicolumn{1}{|c|}{Model} & \multicolumn{1}{c|}{Params (M) } & \multicolumn{1}{c|}{CFG} & \multicolumn{1}{c|}{FID - 50K $\downarrow$ } & \multicolumn{1}{c|}{SFID $\downarrow$} & \multicolumn{1}{c|}{IS $\uparrow$} & \multicolumn{1}{c|}{Precision $\uparrow$} & Recall $\uparrow$  \\ \hline
\multicolumn{1}{|c|}{DiT - XL/2 \cite{peebles2023dit}} & \multicolumn{1}{c|}{675} & \multicolumn{1}{c|}{4} & \multicolumn{1}{c|}{16.23} & \multicolumn{1}{c|}{11.06} & \multicolumn{1}{c|}{80.91} & \multicolumn{1}{c|}{0.93} & 0.26 \\ \hline
\multicolumn{1}{|c|}{EdgeDiT 1} & \multicolumn{1}{c|}{471} & \multicolumn{1}{c|}{4} & \multicolumn{1}{c|}{12.3} & \multicolumn{1}{c|}{13.97} & \multicolumn{1}{c|}{75.72} & \multicolumn{1}{c|}{0.92} & 0.24 \\ \hline
\multicolumn{1}{|c|}{EdgeDiT 6} & \multicolumn{1}{c|}{530} & \multicolumn{1}{c|}{4} & \multicolumn{1}{c|}{12.4} & \multicolumn{1}{c|}{14.96} & \multicolumn{1}{c|}{78} & \multicolumn{1}{c|}{0.91} & 0.25 \\ \hline
\end{tabular}%
}
\end{table*}

We evaluated this by replacing the MLP ratio in DiT Blocks from 4 to {1,2,3}. Figure \ref{fig:mlp_ratio} shows that image quality drastically decreases when the MLP ratio is set to 1, while MLP ratios 2, 3 produce similar results. So, we select surrogates with MLP ratio 2 in the  search space. 

We have also experimented by replacing the hidden dimension in DiT blocks to 768 instead of 1152. Figure \ref{fig:att_dim} shows that image quality is very similar to that generated by the surrogate having hidden dimension as 512. Due to similar image quality, we include surrogates with hidden dimension 512 to the search space. These combined efforts brought down the search space from $6^{28}$ to $4^{28}$ in Stage 2.

\subsection{Pareto Analysis and Model Selection}
To jointly evaluate generative quality and deployment efficiency, we formulate model selection as a bi-objective optimization problem with two competing objectives: (i) minimizing the Fréchet Inception Distance (FID), which measures the fidelity of generated images, and (ii) minimizing the on-device latency loss, which captures the inference overhead on the target edge hardware. 

Since improvements in generative quality often come at the cost of increased computational complexity and runtime, these objectives naturally exhibit a trade-off. We therefore analyze the resulting candidate models using the Pareto front, which represents the set of non-dominated solutions where no model can improve one objective without degrading the other. Models lying on this front provide optimal trade-offs between visual quality and inference efficiency. 

From the resulting Pareto set, we select two representative architectures for full training and evaluation: \textit{EdgeDiT-1} and \textit{EdgeDiT-6}. These models correspond to the smallest and largest architectures within the EdgeDiT family in terms of parameter count, thereby capturing the spectrum of efficiency–capacity trade-offs discovered by the search. Due to limited computational resources, it was not feasible to train every candidate architecture identified during the search. Instead, we employ a 400K iteration checkpoint of DiT-XL/2 \cite{peebles2023dit} as the teacher model for surrogate guidance and also treat it as the baseline architecture, as it is the only publicly available checkpoint within the DiT family. 

The selected EdgeDiT-1 and EdgeDiT-6 models are then trained end-to-end with an additional 100K iterations for class conditioned image generation task on ImageNet \cite{deng2009imagenet} dataset to obtain their final weights. All training experiments were conducted using \textbf{4 NVIDIA A100 GPUs}.

\subsection{Accuracy and Visual Comparisons}

We compared the EdgeDiT 1, 6 models with baseline DiT-XL/2 \cite{peebles2023dit} architecture both qualitatively and quantitatively. Quantitative comparisons are reported in Table \ref{tab:fid_results}. Table \ref{tab:fid_results} shows that with minimal training of the EdgeDiT candidate architectures, We were able to outperform the baseline DiT-XL/2 \cite{peebles2023dit} architecture. Although EdgeDiT has around $20-30$\% reduction in parameters than the baseline architecture, it was able to outperform the baseline architecture with minimal training, this demonstrates efficient surrogate selection while preserving both quality and efficiency.

Visual results are shown in  Figure \ref{fig:visual_results}, Figure \ref{fig:visual_results} shows that the quality of generated images from EdgeDiT is comparable or even better than DiT-XL/2 \cite{peebles2023dit} in some cases of class-conditioned image generation.

% \subsection{Server Benchmarking}
\subsection{Performance Comparisons}
\begin{itemize}

 \item \textbf{Server Benchmarking:} We have calculated the parameters, GFLOPs, GMACs of the DiT family of models (S, B, L, XL) and our EdgeDiT for both (256, 256) and (512, 512) image generation. Results have been reported in Table \ref{tab:server_complexity}
 \item \textbf{Edge Device Benchmarking:} We have profiled the DiT family of models (S, B, L, XL) and our EdgeDiT on Apple A18 Pro ANE (iPhone 16 Pro Max) and Qualcomm 8850 NPU (Samsung Galaxy S25 Ultra). Results are presented in Table \ref{tab:device_complexity}.  
\end{itemize}
Tables \ref{tab:server_complexity} and \ref{tab:device_complexity} show that EdgeDiT models had around $\mathbf{20-30}$\% fewer parameters, $\mathbf{28-40}$\% fewer GMACs, and $\mathbf{36-46}$\% reduced GFLOPs than the DiT-XL/2 \cite{peebles2023dit} baselines. While on the edge device latency, EdgeDiT models are $\mathbf{1.65 \times}$ faster on Samsung Galaxy S25 Ultra, and $\mathbf{1.45 \times}$ faster on iPhone 16 Pro Max than the DiT-XL/2 \cite{peebles2023dit} baselines.

\begin{table*}[]
\centering
\caption{Complexity Analysis: Parameters, GMACs, and GFLOPS for varying image resolutions.}
\label{tab:server_complexity}
\resizebox{\textwidth}{!}{%
\begin{tabular}{|c|ccc|ccc|}
\hline
\multirow{2}{*}{Model} & \multicolumn{3}{c|}{Image Size: $1 \times 4 \times 256 \times 256$} & \multicolumn{3}{c|}{Image Size: $1 \times 4 \times 512 \times 512$} \\ \cline{2-7} 
 & \multicolumn{1}{c|}{Params (M)} & \multicolumn{1}{c|}{GMACs} & GFlops & \multicolumn{1}{c|}{Params   (M)} & \multicolumn{1}{c|}{GMACs} & GFlops \\ \hline
DiT S/2 & \multicolumn{1}{c|}{32.96} & \multicolumn{1}{c|}{6.06} & 12.13 & \multicolumn{1}{c|}{33.26} & \multicolumn{1}{c|}{31.44} & 62.93 \\ \hline
DiT B/2 & \multicolumn{1}{c|}{130.51} & \multicolumn{1}{c|}{23.01} & 46.04 & \multicolumn{1}{c|}{131.10} & \multicolumn{1}{c|}{106.38} & 212.88 \\ \hline
DiT L/2 & \multicolumn{1}{c|}{458.10} & \multicolumn{1}{c|}{80.70} & 161.47 & \multicolumn{1}{c|}{458.89} & \multicolumn{1}{c|}{360.98} & 722.27 \\ \hline
DiT XL /2 & \multicolumn{1}{c|}{675.13} & \multicolumn{1}{c|}{118.62} & 237.34 & \multicolumn{1}{c|}{676.01} & \multicolumn{1}{c|}{524.54} & 1050.00 \\ \hline
EdgeDiT 1 & \multicolumn{1}{c|}{470.65} & \multicolumn{1}{c|}{71.94} & 143.96 & \multicolumn{1}{c|}{470.76} & \multicolumn{1}{c|}{311.29} & 560.01 \\ \hline
EdgeDiT 2 & \multicolumn{1}{c|}{480.01} & \multicolumn{1}{c|}{74.02} & 148.12 & \multicolumn{1}{c|}{480.97} & \multicolumn{1}{c|}{336.74} & 605.93 \\ \hline
EdgeDiT 3 & \multicolumn{1}{c|}{491.88} & \multicolumn{1}{c|}{76.62} & 153.32 & \multicolumn{1}{c|}{492.77} & \multicolumn{1}{c|}{342.11} & 684.55 \\ \hline
EdgeDiT 4 & \multicolumn{1}{c|}{501.32} & \multicolumn{1}{c|}{78.70} & 157.49 & \multicolumn{1}{c|}{502.21} & \multicolumn{1}{c|}{346.41} & 693.15 \\ \hline
EdgeDiT 5 & \multicolumn{1}{c|}{510.76} & \multicolumn{1}{c|}{80.78} & 161.65 & \multicolumn{1}{c|}{511.75} & \multicolumn{1}{c|}{350.70} & 711.65 \\ \hline
EdgeDiT 6 & \multicolumn{1}{c|}{529.64} & \multicolumn{1}{c|}{84.94} & 169.97 & \multicolumn{1}{c|}{529.69} & \multicolumn{1}{c|}{372.86} & 762.83 \\ \hline
\end{tabular}%
}
\end{table*}

\begin{table*}[]
\centering
\caption{Inference Latency (ms) comparison on mobile devices across different resolutions.}
\label{tab:device_complexity}
\resizebox{\textwidth}{!}{%
\begin{tabular}{|c|cc|cc|}
\hline
\multirow{2}{*}{Model} & \multicolumn{2}{c|}{Image Size: $1 \times 4 \times 256 \times 256$} & \multicolumn{2}{c|}{Image Size: $1 \times 4 \times 512 \times 512$} \\ \cline{2-5} 
 & \multicolumn{1}{c|}{iPhone Latency} & Samsung   Latency & \multicolumn{1}{c|}{iPhone   Latency} & Samsung   Latency \\ \hline
DiT S/2 & \multicolumn{1}{c|}{6.63} & 8.35 & \multicolumn{1}{c|}{93.31} & 52 \\ \hline
DiT B/2 & \multicolumn{1}{c|}{19.55} & 24.45 & \multicolumn{1}{c|}{286.58} & 115.95 \\ \hline
DiT L/2 & \multicolumn{1}{c|}{68.99} & 85.00 & \multicolumn{1}{c|}{790.26} & 376.08 \\ \hline
DiT XL /2 & \multicolumn{1}{c|}{118.56} & 129.00 & \multicolumn{1}{c|}{1098.92} & 553.13 \\ \hline
EdgeDiT 1 & \multicolumn{1}{c|}{70.86} & 86.13 & \multicolumn{1}{c|}{804.81} & 379.22 \\ \hline
EdgeDiT 2 & \multicolumn{1}{c|}{70.89} & 86.55 & \multicolumn{1}{c|}{808.67} & 381.5 \\ \hline
EdgeDiT 3 & \multicolumn{1}{c|}{71.35} & 88.19 & \multicolumn{1}{c|}{810.17} & 383.69 \\ \hline
EdgeDiT 4 & \multicolumn{1}{c|}{72.36} & 87.81 & \multicolumn{1}{c|}{816.59} & 385.42 \\ \hline
EdgeDiT 5 & \multicolumn{1}{c|}{71.22} & 88.74 & \multicolumn{1}{c|}{811.29} & 381.56 \\ \hline
EdgeDiT 6 & \multicolumn{1}{c|}{72.53} & 89.22 & \multicolumn{1}{c|}{820.37} & 389.37 \\ \hline
\end{tabular}%
}
\end{table*}

\subsection{Ablation Analysis}
\textbf{Impact of feature-wise Knowledge Distillation:} FwKD is central to our EdgeDiT design. It reduces the computational cost required for end-to-end training of candidate architectures. To validate its contribution, we perform an ablation in which EdgeDiT architectures are initialized with random weights and then trained with progressive learning. We observed that image quality degraded significantly more than the corresponding EdgeDiT models with FwKD. Refer to Figure \ref{fig:fwkd}. These results confirm that FwKD is necessary to achieve the desired accuracy, and it also reduces the computational cost required for training.
\section{Conclusion}
% \label{sec:conclusion}
EdgeDiT demonstrates that the scalability of Diffusion Transformers is not inherently at odds with edge deployment. By leveraging layer-wise distillation and Bayesian-optimized architecture search, we successfully compressed the state-of-the-art DiT-XL/2 by $\mathbf{30}$\%, and improved edge latency by $\mathbf{1.65\times}$. Our approach provides a robust framework for bringing large-scale generative models to mobile NPUs, ensuring privacy, speed, and efficiency for the next generation of mobile AI.

\section{Limitations and Future Work}

While EdgeDiT demonstrates promising results for enabling diffusion transformers on mobile hardware, several limitations remain. First, our experiments are primarily based on the DiT-XL/2\cite{peebles2023dit} architecture as the teacher model. Although the proposed surrogate-based search framework is model-agnostic, extending it to other diffusion transformer families such as SiT\cite{siT2024} or MDT\cite{gao2023mdt} remains future work. 

Second, the objective of this work is not to design the most parameter-efficient diffusion model, but to introduce a general hardware-aware optimization framework capable of discovering efficient architectures within a given model family. Integrating this framework with newer transformer-based diffusion models could further improve performance–efficiency trade-offs.

Third, although we evaluate latency on representative mobile NPUs such as Qualcomm Hexagon and Apple Neural Engine, the search space can be further specialized for different hardware accelerators such as TPUs and other edge devices such as AR/VR/XR.

Finally, due to the high cost of diffusion training, only a subset of candidate architectures is trained end-to-end. Larger compute budgets could enable broader exploration of the architecture space and potentially yield even more efficient designs.
{
    \small
    \bibliographystyle{ieeenat_fullname}
    \bibliography{main}

@String(CVPR= {IEEE Conf. Comput. Vis. Pattern Recog.})

@String(ICCV= {Int. Conf. Comput. Vis.})

@String(CVPR  = {CVPR})

@String(ICCV  = {ICCV})

@inproceedings{ho2020ddpm,
  title={Denoising Diffusion Probabilistic Models},
  author={Ho, Jonathan and Jain, Ajay and Abbeel, Pieter},
  booktitle={NeurIPS},
  year={2020}
}

@article{song2021ddim,
  title={Denoising diffusion implicit models},
  author={Song, Jiaming and Meng, Chenlin and Ermon, Stefano},
  journal={arXiv preprint arXiv:2010.02502},
  year={2020}
}

@inproceedings{nichol2021improved,
  title={Improved Denoising Diffusion Probabilistic Models},
  author={Nichol, Alex and Dhariwal, Prafulla},
  booktitle={ICML},
  year={2021}
}

@inproceedings{rombach2022ldm,
  title={High-Resolution Image Synthesis with Latent Diffusion Models},
  author={Rombach, Robin and Blattmann, Andreas and Lorenz, Dominik and Esser, Patrick and Ommer, Bjorn},
  booktitle={CVPR},
  year={2022}
}

@inproceedings{peebles2023dit,
  title={Scalable Diffusion Models with Transformers},
  author={Peebles, William and Xie, Saining},
  booktitle={ICCV},
  year={2023}
}

@article{hinton2015distilling,
  title={Distilling the knowledge in a neural network},
  author={Hinton, Geoffrey and Vinyals, Oriol and Dean, Jeff},
  journal={arXiv preprint arXiv:1503.02531},
  year={2015}
}

@article{snoek2012practical,
  title={Practical bayesian optimization of machine learning algorithms},
  author={Snoek, Jasper and Larochelle, Hugo and Adams, Ryan P},
  journal={Advances in neural information processing systems},
  volume={25},
  year={2012}
}

@inproceedings{mobilediffusion,
  title={Mobilediffusion: Instant text-to-image generation on mobile devices},
  author={Zhao, Yang and Xu, Yanwu and Xiao, Zhisheng and Jia, Haolin and Hou, Tingbo},
  booktitle={European Conference on Computer Vision},
  pages={225--242},
  year={2024},
  organization={Springer}
}

@inproceedings{siT2024,
  title={Sit: Exploring flow and diffusion-based generative models with scalable interpolant transformers},
  author={Ma, Nanye and Goldstein, Mark and Albergo, Michael S and Boffi, Nicholas M and Vanden-Eijnden, Eric and Xie, Saining},
  booktitle={European Conference on Computer Vision},
  pages={23--40},
  year={2024},
  organization={Springer}
}

@article{statespace2024diffusion,
  title={Scalable Diffusion Models with State Space Backbone. arXiv 2024},
  author={Fei, Z and Fan, M and Yu, C and Huang, J},
  journal={arXiv preprint arXiv:2402.05608},
  year={2024}
}

@inproceedings{gao2023mdt,
  title={Masked diffusion transformer is a strong image synthesizer},
  author={Gao, Shanghua and Zhou, Pan and Cheng, Ming-Ming and Yan, Shuicheng},
  booktitle={Proceedings of the IEEE/CVF international conference on computer vision},
  pages={23164--23173},
  year={2023}
}

@article{gao2023mdtv2,
  title={Mdtv2: Masked diffusion transformer is a strong image synthesizer},
  author={Gao, Shanghua and Zhou, Pan and Cheng, Ming-Ming and Yan, Shuicheng},
  journal={arXiv preprint arXiv:2303.14389},
  year={2023}
}

@article{li2024representation,
  title={Representation alignment for generation: Training diffusion transformers is easier than you think},
  author={Yu, Sihyun and Kwak, Sangkyung and Jang, Huiwon and Jeong, Jongheon and Huang, Jonathan and Shin, Jinwoo and Xie, Saining},
  journal={arXiv preprint arXiv:2410.06940},
  year={2024}
}

@inproceedings{deng2009imagenet,
  title={Imagenet: A large-scale hierarchical image database},
  author={Deng, Jia and Dong, Wei and Socher, Richard and Li, Li-Jia and Li, Kai and Fei-Fei, Li},
  booktitle={2009 IEEE conference on computer vision and pattern recognition},
  pages={248--255},
  year={2009},
  organization={Ieee}
}
}


\end{document}